# Enhancing Breast Cancer Prediction with LLM-Inferred Confounders

Debmita Roy
Wheeler High School, Marietta, GA

**Abstract**
*This study enhances breast cancer prediction by using large language models to infer the likelihood of confounding diseases, namely diabetes, obesity, and cardiovascular disease, from routine clinical data. These AI-generated features improved Random Forest model performance, particularly for LLMs like Gemma (3.9%) and Llama (6.4%). The approach shows promise for noninvasive prescreening and clinical integration, supporting improved early detection and shared decision-making in breast cancer diagnosis.*

**Introduction**
Breast cancer (BC) is a leading cause of death among women in the U.S., with most cases having unknown causes despite known risk factors[1]. Researchers have identified correlations between BC and various clinical features and biomarkers, such as body mass index, glucose, insulin, leptin, adiponectin, resistin, MCP-1, and HOMA, that can be measured through routine blood tests. Machine learning (ML) models using these features have been developed to support BC prediction and screening[2]. However, understanding confounding diseases is critical when using these metabolic and inflammatory biomarkers to predict BC, because many other diseases can alter these markers, reducing specificity. A literature review identified Type-2 Diabetes, Obesity, and Cardiovascular Disease (CVD) as key confounders[3,4]. In this paper, we propose to leverage Large Language Models (LLMs) to generate proxy measures for confounders not included in the original data and to integrate them with clinical features to enhance BC prediction models. This approach aims to mitigate confounding effects in prescreening, providing oncologists with comprehensive insights to support BC diagnosis.

**Methods**
This study used the Breast Cancer Coimbra dataset, which includes the above nine clinical features measured for 64 patients with BC and 52 healthy controls[2]. From the Together AI platform, we selected two LLMs with different sizes, Llama-3.3-70B (Meta, 70 billion parameters)[5] and Gemma-2-27B (Google, 27 billion parameters)[6], for inference via API service. We prompted both models with each patient's clinical features to estimate the likelihood (probability scores between 0 and 1) of having three confounding diseases: Diabetes, Obesity, and CVD. As a benchmark, we also used the models to predict the likelihood of breast cancer itself. These four AI-generated synthetic features were augmented to the original nine features. To evaluate model performance, we randomly split the data into 20 different 80/20 train-test pairs. For each split, we trained Random Forest classifiers under several feature configurations: (1) *Baseline* - Original nine clinical features; (2) *LLM-only* - LLM-generated breast cancer likelihood; (3) *Single-confounder* - Nine features with one confounder likelihood (Diabetes, Obesity, or CVD); and (4) *All-confounders* - Nine features with all three confounder likelihoods. Performance metrics (accuracy, precision, recall, AUC, and specificity) were averaged across all 20 iterations to obtain more stable and less biased estimates of the model's performance. Standard errors indicated variability, and the pipeline was run separately for both Llama and Gemma, but it is compatible with any LLM accessible via API.

**Results**
The Random Forest classifier's performance across feature configurations using both LLMs is summarized in Table 1. As expected, the LLM-only setup performed the weakest, while adding a single confounder consistently outperformed the baseline for both models. The wAllConfounders configuration produced the highest gains, improving over the baseline by an average of 3.9% (Gemma) and 6.4% (Llama). Figure 1 illustrates that Llama generally outperformed Gemma in this setup, except on recall—likely due to its larger model size and stronger ability for meta-learning. Overall, incorporating AI-generated synthetic features improved predictions above baseline, highlighting their potential to reduce confounder effects and enhance clinical integration.

**Discussion and Conclusions**
This study shows that LLM-generated confounder likelihoods can improve BC prediction using routine clinical data. Augmenting features with inferred risks of diabetes, obesity, and CVD boosted model performance, particularly with larger models like Llama. Despite promising results, limitations include a small, homogeneous dataset and a lack of access to more powerful LLMs. Future work should involve larger, more diverse cohorts, additional confounders (e.g., Polycystic Ovary Syndrome[7], autoimmune diseases), additional systemic conditions (e.g., anemia)[8], and incorporating interpretability by generating explanations for confounder likelihoods for enhancing clinical trust. Overall, this approach shows promise for improving early breast cancer screening and supporting Human-AI shared decision-making, especially when integrated into electronic health record systems.

**Table 1.** Performance metrics of Random Forest classifier under feature configurations: (1) Baseline - Original nine clinical features; (2) LLM-only - LLM-generated breast cancer likelihood; (3) Single-confounder - Nine features with one confounder likelihood (Diabetes, Obesity, or CVD); and (4) All-confounders - Nine features with all three confounder likelihoods. The performance corresponding to the LLMs, Gemma and Llama, is displayed separately.

| | Baseline Classifier | Gemma-2-27B | | | | | Baseline Classifier | Llama-3.3-70B | | | | |
|---|---|---|---|---|---|---|---|---|---|---|---|---|
| | | LLM | wDiabetes | wCVD | wObesity | wAllConfounders | | LLM | wDiabetes | wCVD | wObesity | wAllConfounders |
| **Accuracy** | 0.719 | 0.645 | 0.725 | 0.731 | 0.733 | **0.746** | 0.704 | 0.674 | 0.712 | 0.719 | 0.715 | **0.750** |
| **Precision** | 0.718 | 0.695 | 0.720 | 0.723 | 0.729 | **0.742** | 0.692 | 0.668 | 0.695 | 0.705 | 0.702 | **0.749** |
| **Recall** | 0.619 | 0.437 | 0.630 | 0.644 | 0.636 | **0.667** | 0.674 | 0.594 | 0.661 | 0.672 | 0.661 | **0.683** |
| **AUC** | 0.710 | 0.626 | 0.715 | 0.721 | 0.725 | **0.740** | 0.711 | 0.664 | 0.716 | 0.721 | 0.716 | **0.753** |
| **Specificity** | 0.801 | 0.814 | 0.801 | 0.798 | 0.813 | **0.813** | 0.749 | 0.734 | 0.770 | 0.770 | 0.771 | **0.822** |

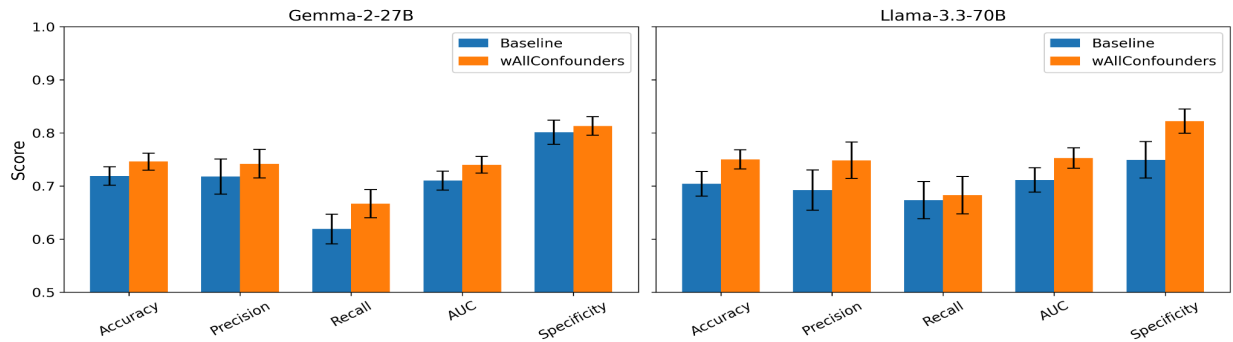

**Figure 1.** Performance comparison with error bars for a baseline Random Forest (RF) classifier vs a RF classifier trained in the All-confounders configuration. In terms of percentage improvement over baseline, Llama consistently outperformed Gemma in the All-confounders setup across all metrics, except recall.

**Acknowledgments**

The author would like to thank Dr. Wang Zhang from the Massachusetts Institute of Technology for his invaluable mentorship throughout this project, including weekly guidance on problem formulation, machine-learning concepts, AI model inference, and experimental iterations. His feedback during the implementation phase and manuscript preparation greatly strengthened the quality of this work. The author also wishes to thank Dr. Subhro Das from the MIT-IBM Watson AI Lab for his mentorship and for inspiring a deeper interest in research in this field.

**References**

1. Obeagu EI, Obeagu GU. Breast cancer: A review of risk factors and diagnosis. Medicine. 2024 Jan 19;103(3):e36905.
2. Patrício M, Pereira J, Crisóstomo J, Matafome P, Gomes M, Seiça R, Caramelo F. Using resistin, glucose, age and BMI to predict the presence of breast cancer. BMC Cancer. 2018 Jan 4;18(1):29.
3. Tilg H, Moschen AR. Adipocytokines: Mediators linking adipose tissue, inflammation and immunity. Nature reviews immunology. 2006 Oct 1;6(10):772-83.
4. Cao H. Adipocytokines in obesity and metabolic disease. Journal of Endocrinology. 2014 Feb 1;220(2):T47-59.
5. Dubey A, Jauhri A, Pandey A, Kadian A, Al-Dahle A, Letman A, Mathur A, Schelten A, Yang A, Fan A, Goyal A. The Llama 3 herd of models. arXiv e-prints. 2024 Jul:arXiv-2407.
6. Team G, Riviere M, Pathak S, Sessa PG, Hardin C, Bhupatiraju S, Hussenot L, Mesnard T, Shahriari B, Ramé A, Ferret J. Gemma 2: Improving open language models at a practical size. arXiv preprint arXiv:2408.00118. 2024 Jul 31.
7. Hemati A, Amini L, Haghani S, Hashemi EA. Investigation of the relationship between breast cancer and clinical symptoms of polycystic ovarian syndrome: A case-control study. BMC Women's Health. 2024 Feb;24(1):586. https://doi.org/10.1186/s12905-024-03421-4.
8. Chen X, Zhou H, Lv J. The importance of hypoxia-related to hemoglobin concentration in breast cancer. Cell Biochemistry and Biophysics. 2024 Sep;82(3):1893-906.